\documentclass[11pt]{article}

\usepackage{acl2012}
\makeatletter
\newcommand{\@BIBLABEL}{\@emptybiblabel}
\newcommand{\@emptybiblabel}[1]{}
\makeatother
\usepackage[hidelinks]{hyperref}
\usepackage{booktabs}       
\usepackage{amsfonts}       
\usepackage{nicefrac}       
\usepackage{microtype}      
\usepackage{graphicx}
\usepackage{titlesec}
\usepackage{subfig}
\usepackage{tikz}
\usepackage{tikz-qtree}
\usepackage{tikz-dependency}
\usepackage{floatrow}
\usepackage{linguex}
\usepackage{color}
\usepackage{times}

\newcommand{\depmem}[1] {\textbf{#1}}
\newcommand{\consistent}[1] {\textit{#1}}
\newcommand{\attractor}[1] {\underline{#1}}
\newcounter{sauvegardeenumi}

\newcommand{\shorten}[1] {\textit{\textcolor{gray}{#1}}}
\renewcommand{\shorten}[1] {}

\title{Assessing the Ability of LSTMs to Learn Syntax-Sensitive Dependencies}

\author{Tal Linzen\textsuperscript{1,2} \qquad Emmanuel Dupoux\textsuperscript{1}\\
        LSCP\textsuperscript{1} \& IJN\textsuperscript{2}, CNRS,\\
        EHESS and ENS, PSL Research University\\ {\tt \{tal.linzen},\\{\tt emmanuel.dupoux\}@ens.fr}
  \And
  Yoav Goldberg\\
  Computer Science Department\\
  Bar Ilan University\\
  \texttt{yoav.goldberg@gmail.com}
}

\begin{document}

\maketitle

\begin{abstract}
The success of long short-term memory (LSTM) neural networks in language
processing is typically attributed to their ability to capture
long-distance statistical regularities. Linguistic regularities
are often sensitive to syntactic structure; can such 
dependencies be captured by LSTMs, which do not have
explicit structural representations? We begin addressing this question using
number agreement in English subject-verb dependencies. We probe the
architecture's grammatical competence both using training objectives with 
an explicit grammatical target
(number prediction, grammaticality judgments) and using language
models. In the strongly supervised settings, the LSTM achieved very high overall accuracy (less than 1\% errors), but errors increased when sequential and structural information conflicted. The
frequency of such errors rose sharply in the language-modeling
setting.
We conclude that LSTMs can capture a non-trivial amount of grammatical structure
given targeted supervision, but stronger architectures may be
required to further reduce errors;
furthermore, the language modeling signal is insufficient for capturing syntax-sensitive dependencies, and should be
supplemented with more direct supervision if such dependencies need to be captured.
\end{abstract}

\section{Introduction}

Recurrent neural networks (RNNs) are highly effective models of sequential data
\cite{elman1990finding}. The rapid adoption of RNNs in NLP systems in recent
years, in particular of RNNs with gating mechanisms such as long short-term
memory (LSTM) units \cite{hochreiter1997long} or gated recurrent units (GRU)
\cite{cho2014learning}, has led to significant gains in language modeling
\cite{mikolov2010recurrent,sundermeyer2012lstm}, parsing
\cite{vinyals2015grammar,kiperwasser2016bilstm,dyer2016rng}, machine
translation \cite{bahdanau2015neural} and other tasks.

The effectiveness of RNNs\footnote{In this work we use the term RNN to refer to
the entire class of sequential recurrent neural networks. Instances of the
class include long short-term memory networks (LSTM) and the Simple Recurrent
Network (SRN) due to Elman \shortcite{elman1990finding}.} is attributed to
their ability to capture statistical contingencies that may span an arbitrary number of words.  The word
\textit{France}, for example, is more likely to occur somewhere in a sentence that begins
with \textit{Paris} than in a sentence that begins with \textit{Penguins}.  The
fact that an arbitrary number of words can intervene between the mutually
predictive words implies that they cannot be captured by models with a fixed
window such as $n$-gram models, but can in principle be captured by RNNs, which
do not have an architecturally fixed limit on dependency length.

RNNs are sequence models: they do not explicitly incorporate syntactic
structure. Indeed, many word co-occurrence statistics can be captured by treating the
sentence as an unstructured list of words (\textit{Paris}-\textit{France}); it is therefore unsurprising
that RNNs can learn them well. Other dependencies, however, are sensitive to
the syntactic structure of the sentence
\cite{chomsky1965aspects,everaert2015structures}. To what extent can RNNs learn to model such phenomena based only on sequential cues?

Previous research has shown that RNNs (in particular LSTMs) can learn artificial context-free languages \cite{gers2001lstm} as well as nesting and indentation in a programming language \cite{karpathy2016visualizing}. The goal of the present work is to probe their ability to learn \textit{natural language} hierarchical (syntactic) structures from a corpus without syntactic annotations.
As a first step, we focus on a particular dependency that is commonly regarded as evidence for hierarchical structure in human language: English subject-verb agreement, the phenomenon in which
the form of a verb depends on whether the subject is singular or
plural (\textit{the kids play} but \textit{the kid plays}; see additional details in Section \ref{sec:background}). If an RNN-based model succeeded in learning this dependency, that would indicate that it can learn to
approximate or even faithfully implement syntactic structure.

Our main interest is in whether LSTMs have the \textit{capacity} to learn
structural dependencies from a natural corpus. We therefore begin by addressing
this question under the most favorable conditions: training with explicit
supervision. In the setting with the strongest supervision, which we refer to
as the number prediction task, we train it directly on the task of guessing the
number of a verb based on the words that preceded it (Sections
\ref{sec:evaluation} and \ref{sec:results_number_prediction}). We further
experiment with a grammaticality judgment training objective, in which we
provide the model with full sentences annotated as to whether or not they
violate subject-verb number agreement, without an indication of the locus
of the violation (Section \ref{sec:alternative}). Finally, we trained the model
without any grammatical supervision, using a language modeling objective
(predicting the next word). 

Our quantitative results (Section \ref{sec:results_number_prediction}) and
qualitative analysis (Section \ref{sec:error_analysis}) indicate that most
naturally occurring agreement cases in the Wikipedia corpus are easy: they can
be resolved without syntactic information, based only on the sequence of nouns
preceding the verb. This leads to high overall accuracy in all models.  Most of
our experiments focus on the supervised number prediction model.  The accuracy
of this model was lower on harder cases, which require the model to encode or
approximate structural information; nevertheless, it succeeded in recovering
the majority of agreement cases even when four nouns of the opposite number
intervened between the subject and the verb (17\% errors).  Baseline models
failed spectacularly on these hard cases, performing far below chance levels.
Fine-grained analysis revealed that mistakes are much more common when no overt
cues to syntactic structure (in particular function words) are available, as is
the case in noun-noun compounds and reduced relative clauses.  This indicates
that the number prediction model indeed managed to capture a decent amount of
syntactic knowledge, but was overly reliant on function words.

Error rates increased only mildly when we switched to more indirect supervision
consisting only of sentence-level grammaticality annotations without an
indication of the crucial verb.  By contrast, the language model trained
without explicit grammatical supervision performed worse than chance on the harder
agreement prediction cases. Even a state-of-the-art large-scale language model
\cite{jozefowicz2016exploring} was highly sensitive to recent but structurally
irrelevant nouns, making more than five times as many mistakes as the
number prediction model on these harder cases.  These results suggest that explicit
supervision is necessary for learning the agreement dependency using this
architecture, limiting its plausibility as a model of child language
acquisition \cite{elman1990finding}. From a more applied perspective, this result suggests that for
tasks in which it is desirable to capture syntactic dependencies (e.g., machine
translation or language generation), language modeling objectives should be
supplemented by supervision signals that directly capture the desired behavior.

\section{\label{sec:background}Background: Subject-Verb Agreement as Evidence for Syntactic Structure}

The form of an English third-person present tense verb depends on whether the
head of the \textit{syntactic subject} is plural or singular:%
\footnote{
Identifying the head of the subject is typically straightforward.  In what
follows we will use the shorthand ``the subject'' to refer to the head of the
subject.}

\ex.
    \a. The \depmem{key is} on the table.
    \b. *The \depmem{key are} on the table.
    \c. *The \depmem{keys is} on the table.
    \d. The \depmem{keys are} on the table.

While in these examples the subject's head is adjacent to the verb, in general 
the two can be separated by some sentential material:\footnote{In the examples, the
subject and the corresponding verb are marked in boldface,
agreement attractors are underlined and intervening nouns of the same number as
the subject are marked in italics. Asterisks mark unacceptable sentences.}

\ex.
    The \depmem{keys} to the \attractor{cabinet} \depmem{are} on the table.

\begin{figure}[h!]
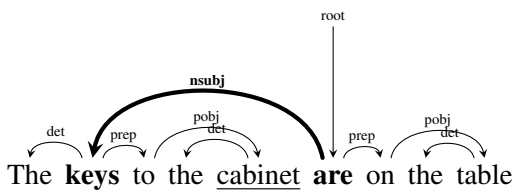
    
    \begin{dependency}[theme=simple]
\begin{deptext}
    The \& \depmem{keys} \& to \& the \& \attractor{cabinet} \& \depmem{are} \& on \& the \& table \\
\end{deptext}
\depedge{2}{1}{det}  
\depedge[edge unit distance=2.2ex,edge style=ultra thick]{6}{2}{\textbf{nsubj}} 
\depedge{2}{3}{prep}   
\depedge{5}{4}{det}   
\depedge{3}{5}{pobj}
\depedge{6}{7}{prep}  
\depedge{9}{8}{det}  
\depedge{7}{9}{pobj}  
\deproot{6}{root}
\end{dependency}

    \caption{The form of the verb is determined by the head of the subject,
    which is directly connected to it via an \textit{nsubj} edge.
    Other nouns that intervene between the
    head of the subject and the verb (here \textit{cabinet} is such a noun) are 
    irrelevant for determining the form of the verb and need
    to be ignored.}
    \label{fig:trees}
\end{figure}

Given a syntactic parse of the sentence and a verb, it is 
straightforward to identify the head of the subject that corresponds to that
verb, and use that information to determine the number of the verb (Figure
\ref{fig:trees}). 

By contrast, models that are insensitive to structure may run into substantial
difficulties capturing this dependency. One potential issue is that there is no
limit to the complexity of the subject NP, and any number of sentence-level
modifiers and parentheticals---and therefore an arbitrary number of words---can
appear between the subject and the verb:

\ex.\label{ex:unbounded} The \depmem{building} on the far right that's quite old and run down \depmem{is} the
 Kilgore Bank Building.

This property of the dependency entails that it cannot be captured by an
$n$-gram model with a fixed $n$. RNNs are in principle able to capture
dependencies of an unbounded length; however, it is an empirical question whether or not they will learn to do so in practice when trained on a natural corpus.

A more fundamental challenge that the dependency poses for structure-insensitive models is the possibility  of \textit{agreement attraction errors} \cite{bock1991broken}. The correct form in \ref{ex:unbounded} could be selected using simple
heuristics such as ``agree with the most recent noun'', which are readily
available to sequence models. In general, however, such heuristics are
unreliable, since other nouns can intervene between the subject and the verb in
the linear sequence of the sentence. Those intervening nouns can have
the same number as the subject, as in \ref{ex:local_consistent}, or the
opposite number as in \ref{ex:local_overt}-\ref{ex:three_attractors}:
 
\ex.\label{ex:local_consistent}Alluvial \depmem{soils} carried in the
\consistent{floodwaters} \depmem{add} nutrients to the floodplains.

\ex.\label{ex:local_overt}The only championship \depmem{banners} that are
currently displayed within the \attractor{building} \depmem{are} for national or NCAA
Championships.

\ex.\label{ex:one_attractor}The \depmem{length} of the \attractor{forewings} \depmem{is} 12-13.

\ex.\label{ex:three_attractors}Yet the \depmem{ratio} of \attractor{men} who survive to the \attractor{women} and \attractor{children} who survive \depmem{is} not clear in this story.

Intervening nouns with the opposite number from the subject are called
\textit{agreement attractors}. The potential
presence of agreement attractor entails that the model must identify the head of the
syntactic subject that corresponds to a given verb in order to choose the correct
inflected form of that verb. 

Given the difficulty in identifying the subject from the linear sequence of the
sentence, dependencies such as subject-verb agreement serve as an argument for
structured syntactic representations in humans \cite{everaert2015structures};
they may challenge models such as RNNs that do not have pre-wired syntactic
representations.  We note that subject-verb number agreement is only one of
a number of structure-sensitive dependencies; other examples include negative
polarity items (e.g., \textit{any}) and reflexive pronouns (\textit{herself}).
Nonetheless, a model's success in learning subject-verb agreement would be
highly suggestive of its ability to master hierarchical structure.

\section{\label{sec:evaluation}The Number Prediction Task}

To what extent can a sequence model learn to be sensitive to the hierarchical
structure of natural language?  To study this question, we propose the
\textit{number prediction} task. In this task, the model sees the sentence up
to but not including a present-tense verb, e.g.: 

 \ex.The keys to the cabinet \hspace{0.1cm}\underline{\hspace{0.8cm}}

\noindent It then needs to guess the number of the following verb (a binary
choice, either \textsc{plural} or \textsc{singular}).  We examine variations on this task
in Section \ref{sec:alternative}.

In order to perform well on this task, the model needs to encode the concepts
of \textit{syntactic number} and \textit{syntactic subjecthood}: it needs to
learn that some words are singular and others are plural, and to be able to
identify the correct subject. As we have illustrated in Section
\ref{sec:background}, correctly identifying the subject that corresponds to a particular
verb often requires sensitivity to hierarchical syntax.

\paragraph{Data:} An appealing property of the number prediction task is that
we can generate practically unlimited training and testing examples for this
task by querying a corpus for sentences with present-tense verbs, and noting
the number of the verb.  Importantly, we do not need to correctly identify the
subject in order to create a training or test example. We generated a corpus of
$\sim$1.35 million number prediction problems based on Wikipedia, of which
$\sim$121,500 (9\%) were used for training, $\sim$13,500 (1\%) for validation,
and the remaining $\sim$1.21 million (90\%) were reserved for
testing.\footnote{We limited our search to sentences that were shorter than 50
words. Whenever a sentence had more than one subject-verb dependency, we
selected one of the dependencies at random.}
The large number of test sentences was necessary to ensure that we had a good
variety of test sentences representing less common constructions
(see Section \ref{sec:results_number_prediction}).\footnote{Code and data
are available at \url{http://tallinzen.net/projects/lstm_agreement}.}

\begin{figure*}
    \raggedright
    \sidesubfloat[]{
        \includegraphics[width=1.8in]{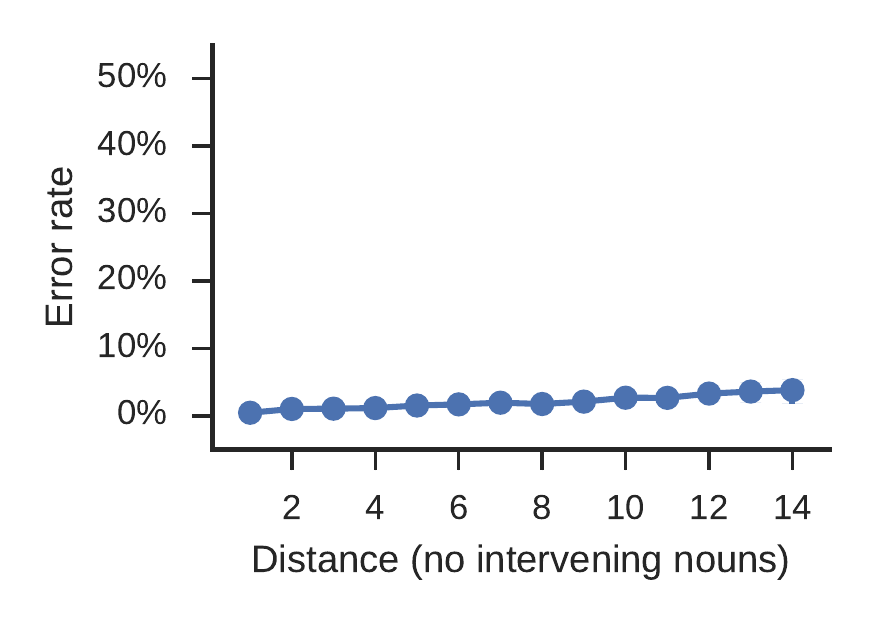}
        \label{fig:np_distance}
    }
    \sidesubfloat[]{
        \includegraphics[width=1.7in]{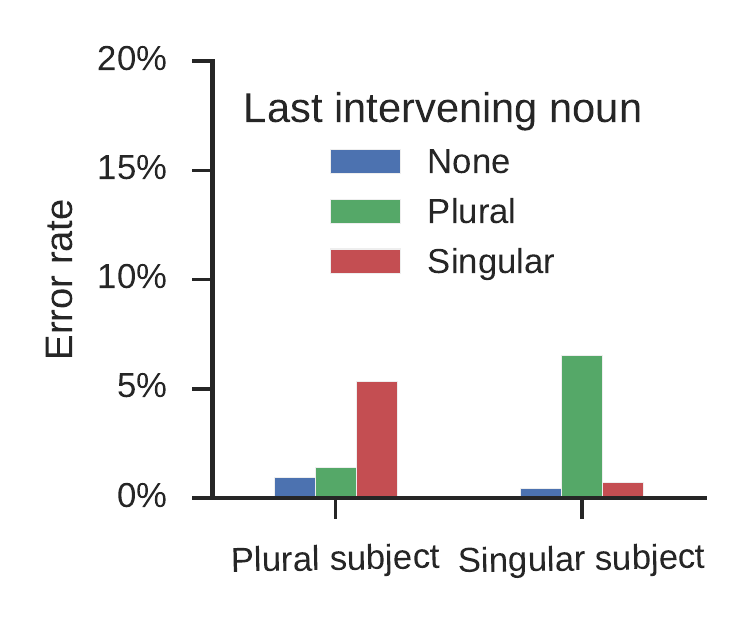}
        \label{fig:np_last_intervening}
    }
     \sidesubfloat[]{
        \includegraphics[width=2in]{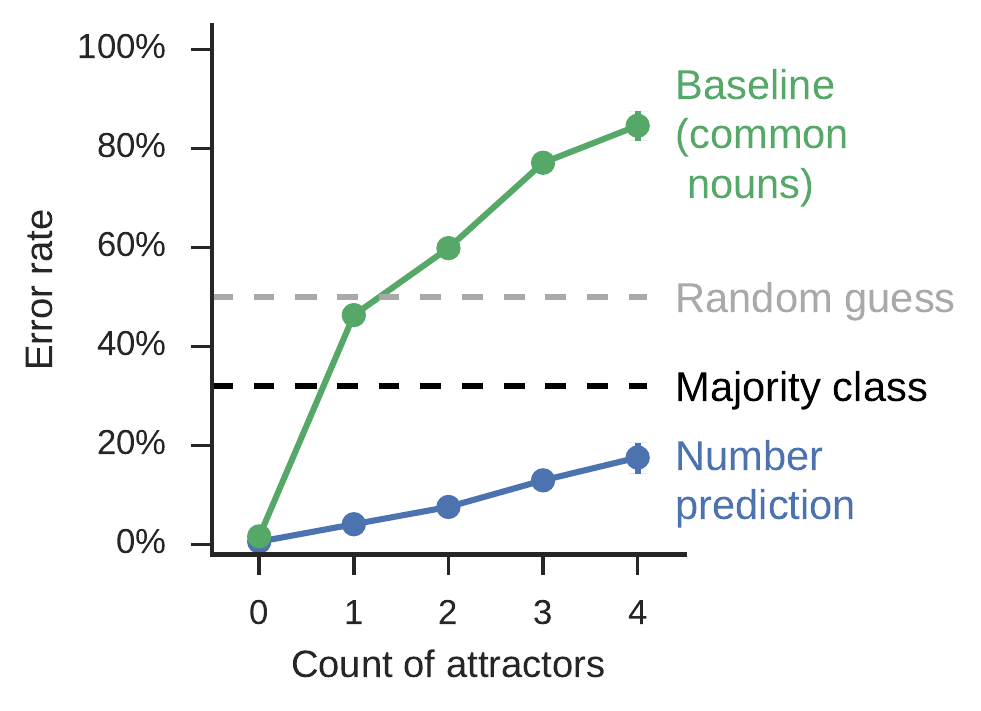}
        \label{fig:np_attractors}
    }

    \sidesubfloat[]{
        \includegraphics[width=1.5in]{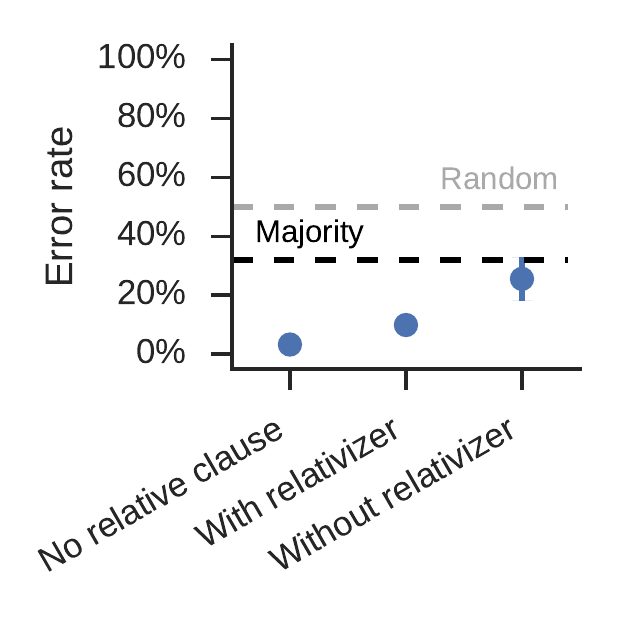}
        \label{fig:np_relativization}
    }
     \sidesubfloat[]{
        \includegraphics[width=1.8in]{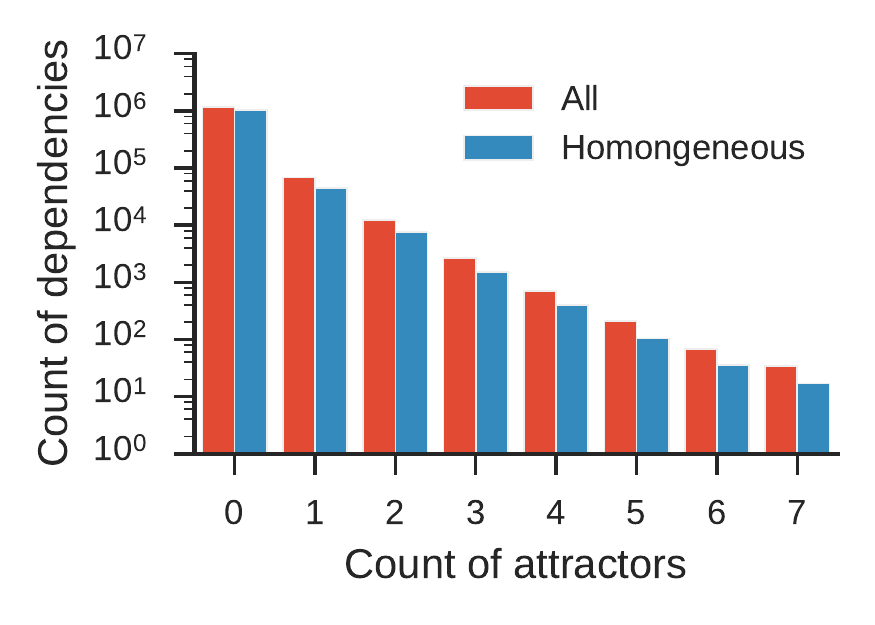}
        \label{fig:n_diff_intervening_count}
    }
    \sidesubfloat[]{
        \includegraphics[width=2in,trim={0 0 0 0},clip]{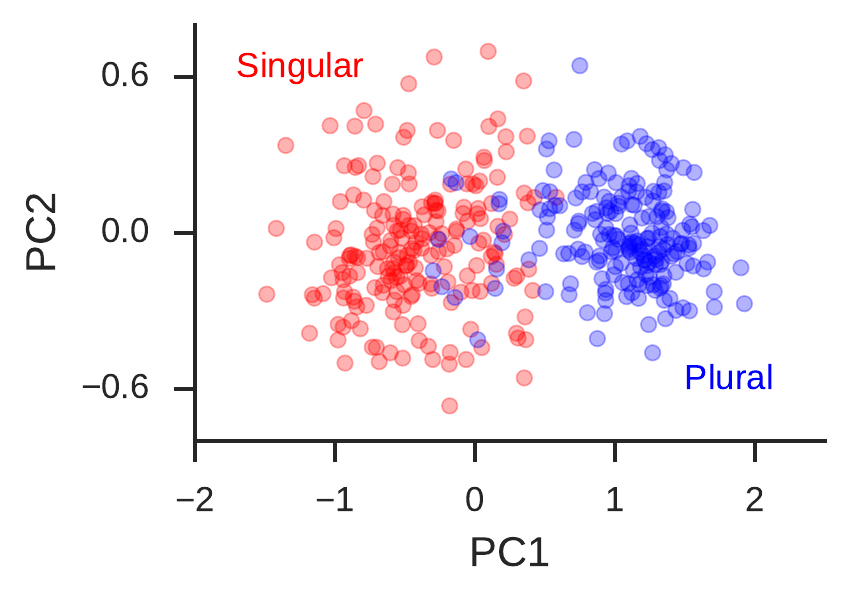}
        \label{fig:embeddings}
    }
    \vspace{-0.4cm}

    \caption[]{\label{fig:np}\textbf{(a-d)} Error rates of the LSTM number prediction model
    as a function of:
    \protect\subref{fig:np_distance} distance between the subject and the
    verb, in dependencies that have no intervening nouns;
    \protect\subref{fig:np_last_intervening} presence and number of last
    intervening noun;   \protect\subref{fig:np_attractors} count of attractors
    in dependencies with homogeneous intervention; \subref{fig:np_relativization} presence of
    a relative clause with and without an overt relativizer in dependencies with homogeneous intervention and exactly
    one attractor. All error bars represent 95\% binomial confidence intervals.
    
    \hspace{0.5cm}\textbf{(e-f)} Additional plots: \protect\subref{fig:n_diff_intervening_count}
    count of attractors per dependency in the corpus (note that the y-axis is
    on a log scale); \protect\subref{fig:embeddings} embeddings of singular and plural nouns, projected onto their first two principal components.} 
    
\end{figure*}

\paragraph{Model and baselines:} We encode words as one-hot vectors: the model
does not have access to the characters that make up the word. Those vectors are
then embedded into a 50-dimensional vector space. An LSTM with 50 hidden units
reads those embedding vectors in sequence; the state of the LSTM at the end of
the sequence is then fed into a logistic regression classifier.
The network is
trained\footnote{The network was optimized using Adam \cite{kingma2015adam} and
early stopping based on validation set error. We trained the number prediction
model 20 times with different random initializations, and report accuracy
averaged across all runs.  The models described in Sections
\ref{sec:alternative} and \ref{sec:additional_experiments} are based on 10
runs, with the exception of the language model, which is slower to train and
was trained once.} in an end-to-end fashion, including the word
embeddings.\footnote{The size of the vocabulary was capped at 10000 (after
lowercasing). Infrequent words were replaced with their part of speech (Penn
Treebank tagset, which explicitly encodes number distinctions); this was the
case for 9.6\% of all tokens and
7.1\% of the subjects. }

To isolate the effect of syntactic structure, we also consider a baseline which
is exposed only to the nouns in the sentence, in the order in which they
appeared originally, and is then asked to predict the
number of the following verb.  The goal of this baseline is to withhold the
syntactic information carried by function words, verbs and other parts of
speech. We explore two variations on this baseline: one that only receives
common nouns (\textit{dogs}, \textit{pipe}), and another that also receives
pronouns (\textit{he}) and proper nouns (\textit{France}). We refer to these as
the \emph{noun-only baselines}.

\section{\label{sec:results_number_prediction}Number Prediction Results}

\paragraph{Overall accuracy:}

Accuracy was very high overall: the system made an incorrect number prediction
only in
0.83\% of the dependencies.  The noun-only baselines performed significantly
  worse: 4.2\% errors for the common-nouns case and 4.5\% errors for the
  all-nouns case. This suggests that function words, verbs and other
  syntactically informative elements play an important role in the model's
  ability to correctly predict the verb's number.  However, while the noun-only
  baselines made more than four times as many mistakes as the number prediction
  system, their still-low absolute error
  rate indicates that around 95\% of agreement dependencies can be
  captured based solely on the sequence of nouns preceding the verb. This is
  perhaps unsurprising: sentences are often short and the verb is often
  directly adjacent to the subject, making the identification of the subject
  simple. To gain deeper insight into the syntactic capabilities of the model, then, the
  rest of this section investigates its performance on more
  challenging dependencies.\footnote{These properties of the
  dependencies were identified by parsing the test sentences using the parser
  described in \newcite{goldberg2012dynamic}.}

\paragraph{Distance:}

We first examine whether the network shows evidence of generalizing to
dependencies where the subject and the verb are far apart. We focus in this
analysis on simpler cases where no nouns intervened between the subject and the
verb. As Figure \ref{fig:np_distance} shows, performance did not degrade
considerably when the distance between the subject and the verb grew up to 15
words (there were very few longer dependencies). This indicates that the
network generalized the dependency from the common distances of 0 and 1 to rare
distances of 10 and more.

\paragraph{Agreement attractors:}

We next examine how the model's error rate was affected by nouns that intervened
between the subject and the verb in the linear order of the sentence. We first
focus on whether or not there were any intervening nouns, and if there were,
whether the number of the subject differed from the number of the last
intervening noun---the type of noun that would trip up the simple
heuristic of agreeing with the most recent noun. 

As Figure \ref{fig:np_last_intervening} shows, a last intervening noun of the same number as the subject
increased error rates only moderately, from 0.4\% to 0.7\% in singular subjects
and from 1\% to 1.4\% in plural subjects. On the other hand, when the last intervening noun was an agreement attractor, error rates increased by almost an order of magnitude (to 6.5\% and 5.4\% respectively).
Note, however, that even an error rate of 6.5\% is quite impressive considering uninformed strategies such as random guessing (50\% error rate), always assigning the more common class label (32\% error rate, since 32\% of the subjects in our corpus are plural) and the number-of-most-recent-noun heuristic (100\% error rate). The noun-only LSTM baselines performed much worse in agreement attraction cases, with error rates of 46.4\% (common nouns) and 40\% (all nouns).

We next tested whether the effect of attractors is cumulative, by focusing on dependencies with multiple attractors. To avoid cases in which the effect
of an attractor is offset by an intervening noun with the same number as the
subject, we restricted our search to dependencies in which all of the intervening nouns had the same number, which we term \textit{dependencies with homogeneous intervention}. For example, \ref{ex:homogeneous_dependency} has homogeneous intervention whereas \ref{ex:non_homogeneous_dependency} does not:

\ex.\label{ex:homogeneous_dependency}The \depmem{roses} in the \attractor{vase} by the \attractor{door} \depmem{are} red.

\ex.\label{ex:non_homogeneous_dependency}The \depmem{roses} in the \attractor{vase} by the \consistent{chairs} \depmem{are} red.

Figure \ref{fig:np_attractors}
shows that error rates increased gradually as more attractors intervened
between the subject and the verb. Performance degraded quite slowly, however: even with four attractors the error
rate was only 17.6\%. As expected, the noun-only baselines performed significantly worse in this setting, reaching an error rate of up to 84\% (worse than chance) in the case of four attractors. This confirms that syntactic cues are critical for solving the harder cases.

\paragraph{Relative clauses:}

We now look in greater detail into the network's performance when the words
that intervened between the subject and verb contained a relative clause. Relative clauses with attractors are likely to be fairly challenging, for several reasons. They typically contain a verb that agrees with the attractor, reinforcing the misleading cue to noun number. The attractor is often itself a subject of an irrelevant verb, making a potential ``agree with the most recent subject'' strategy unreliable. Finally, the existence of a relative clause is sometimes not overtly indicated by a function word (relativizer), as in \ref{ex:covert} (for comparison, see the minimally different \ref{ex:overt}):

\ex.\label{ex:covert} The \depmem{landmarks} this \attractor{article} lists here \depmem{are} also run-of-the-mill and not notable.

\ex.\label{ex:overt} The \depmem{landmarks} \textit{that} this \attractor{article} lists here \depmem{are} also run-of-the-mill and not notable.

For data sparsity reasons we restricted our attention to dependencies with
a single attractor and no other intervening nouns. As Figure \ref{fig:np_relativization} shows, attraction errors were more frequent in dependencies with an overt relative clause (9.9\% errors) than in dependencies without a relative clause (3.2\%), and considerably more frequent when the relative clause was not introduced by an overt relativizer (25\%). As in the case of multiple attractors, however, while the model struggled with the more difficult dependencies, its performance was much better than random guessing, and slightly better than a majority-class strategy.

\paragraph{Word representations:}

We explored the 50-dimensional word representations acquired by the model by
performing a principal component analysis.
We assigned a part-of-speech (POS) to each word based on the
word's most common POS in the corpus. We only considered relatively ambiguous words, in which a single POS accounted
for more than 90\% of the word's occurrences in the corpus.
Figure \ref{fig:embeddings} shows that the first principal component
corresponded almost perfectly to the expected number of the noun, suggesting
that the model learned the number of specific words very well; recall that the
model did not have access during training to noun number annotations or to
morphological suffixes such as \textit{-s} that could be used to identify
plurals.

\begin{figure*}[!ht]
    \sidesubfloat[]{\includegraphics[width=0.55\linewidth]{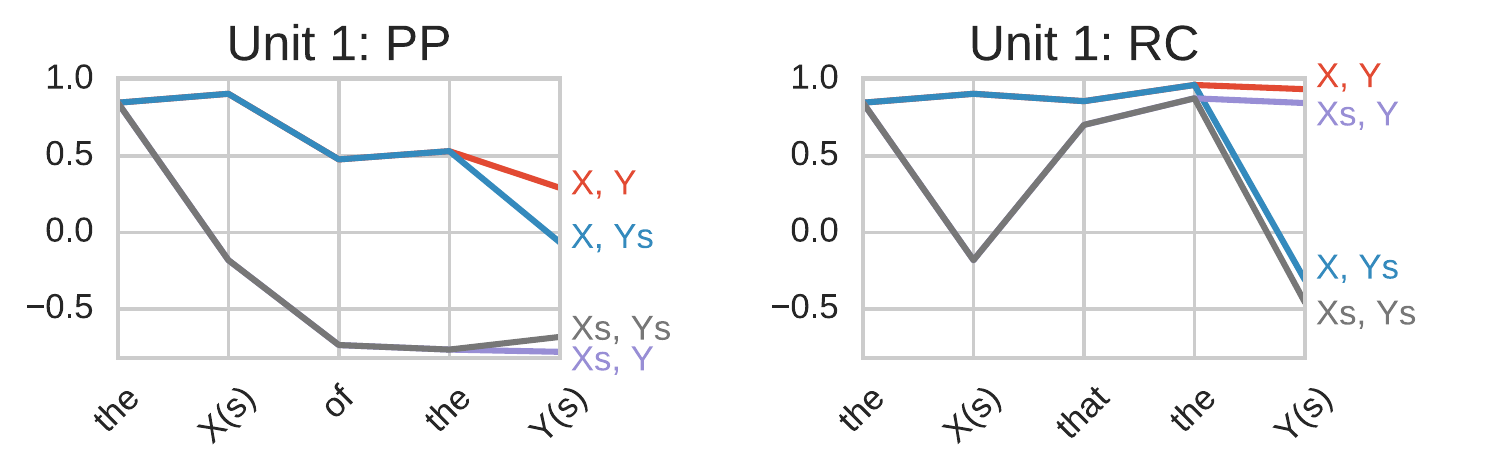}
    \label{fig:lstmviz_factorial}
    }%
    \sidesubfloat[]{
    \includegraphics[width=0.35\linewidth]{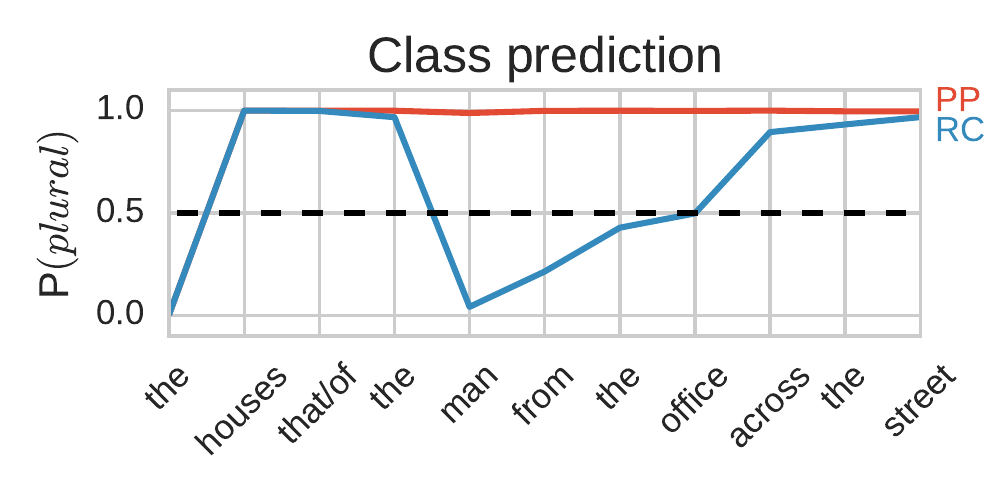}
    \label{fig:lstmviz_predictions}
}
    
    \sidesubfloat[]{\includegraphics[width=0.7\linewidth]{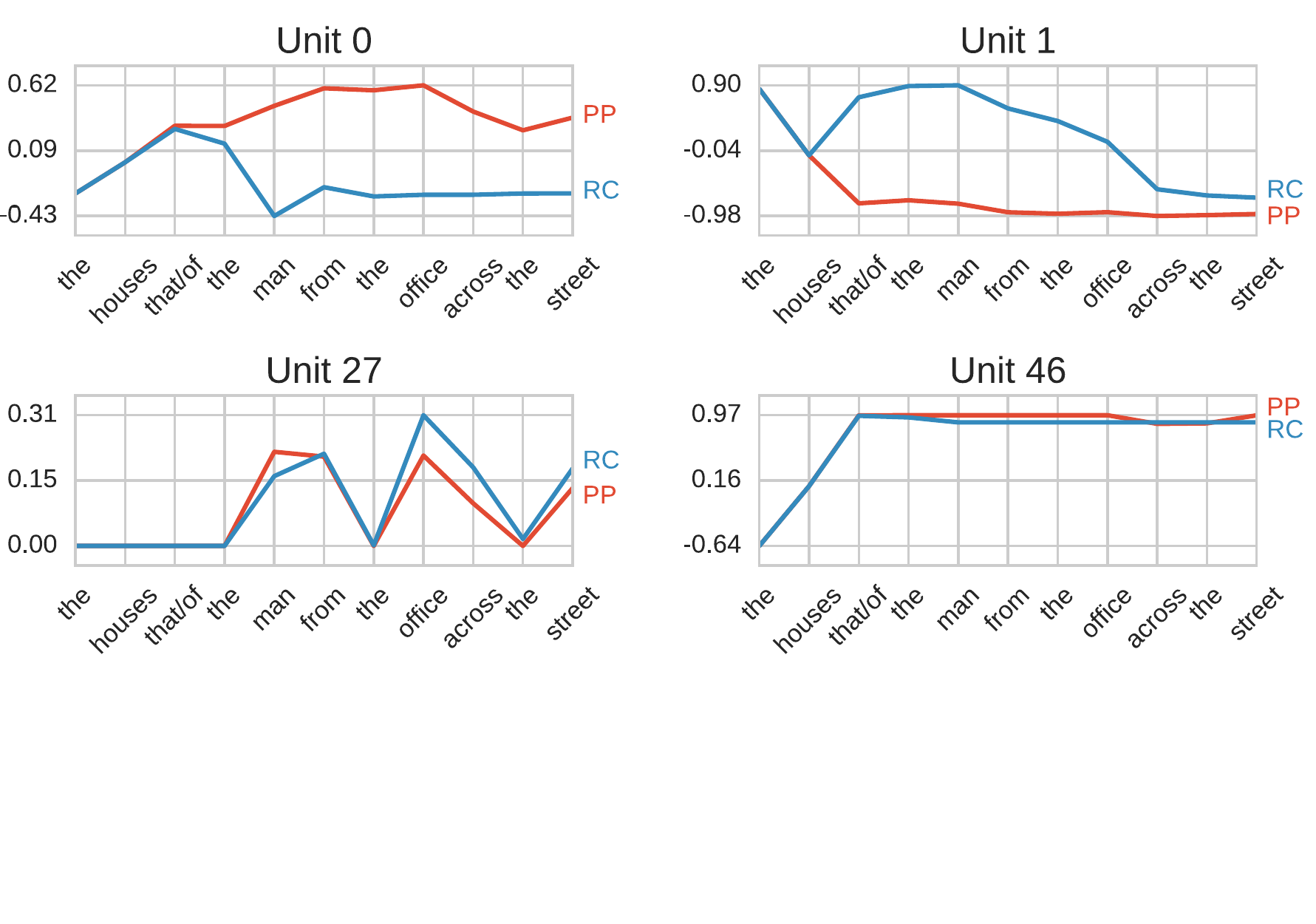}
\label{fig:lstmviz_units}}

    \vspace{-1in}
\caption{Word-by-word visualization of LSTM activation: \protect\subref{fig:lstmviz_factorial} a unit that correctly predicts the number of an upcoming verb. This number is determined by the first noun (X) when the modifier is a prepositional phrase (PP) and by the second noun (Y) when it is an object relative clause (RC); \protect\subref{fig:lstmviz_predictions} the evolution of the predictions in the case of a longer modifier: the predictions correctly diverge at the embedded noun, but then incorrectly converge again; \protect\subref{fig:lstmviz_units} the activation of four representative units over the course of the same sentences.}
\end{figure*}

\paragraph{Visualizing the network's activations:}

We start investigating the inner workings of the number prediction network by analyzing its activation in response to particular syntactic constructions. To simplify the analysis, we deviate from our practice in the rest of this paper and use constructed sentences.

We first constructed sets of sentence prefixes based on the following patterns:

\ex.\makebox[0pt][l]{\textbf{PP:}}\phantom{\textbf{RC:}} The toy(s) of the boy(s)...

\ex.\textbf{RC:} The toy(s) that the boy(s)...
    
These patterns differ by exactly one function word, which determines the type of the modifier of the main clause subject: a prepositional phrase (PP) in the first sentence and a relative clause (RC) in the second. In PP sentences the correct number of the upcoming verb is determined by the main clause subject \textit{toy(s)}; in RC sentences it is determined by the embedded subject \textit{boy(s)}. 

We generated all four versions of each pattern, and repeated the process ten times with different lexical items (\textit{the house(s) of/that the girl(s)}, \textit{the computer(s) of/that the student(s)}, etc.), for a total of 80 sentences. The network made correct number predictions for all 40 PP sentences, but made three errors in RC sentences. We averaged the word-by-word activations across all sets of ten sentences that had the same combination of modifier (PP or RC), first noun number and second noun number. Plots of the activation of all 50 units are provided in the Appendix (Figure \ref{supplemental}). Figure \ref{fig:lstmviz_factorial} highlights a unit (Unit 1) that shows a particularly clear pattern: it tracks the number of the main clause subject throughout the PP modifier, resets when it reaches the relativizer \textit{that} which introduces the RC modifier, and then switches to tracking the number of the embedded subject.

To explore how the network deals with dependencies spanning a larger number of words, we tracked its activation during the processing of the following two sentences:\footnote{We simplified this experiment in light of the relative robustness of the first experiment to lexical items and to whether each of the nouns was singular or plural.}

\ex.The houses of/that the man from the office across the street...

The network made the correct prediction for the PP but not the RC sentence (as before, the correct predictions are \textsc{plural} for PP and \textsc{singular} for RC). Figure \ref{fig:lstmviz_predictions} shows that the network begins by making the correct prediction for RC immediately after \textit{that}, but then falters: as the sentence goes on, the resetting effect of \textit{that} diminishes. The activation time courses shown in Figure \ref{fig:lstmviz_units} illustrate that Unit 1, which identified the subject correctly when the prefix was short, gradually forgets that it is in an embedded clause as the prefix grows longer. By contrast, Unit 0 shows a stable capacity to remember the current embedding status. Additional representative units shown in Figure \ref{fig:lstmviz_units} are Unit 46, which consistently stores the number of the main clause subject, and Unit 27, which tracks the number of the most recent noun, resetting at noun phrase boundaries.

While the interpretability of these patterns is encouraging, our analysis only scratches the surface of the rich possibilities of a linguistically-informed analysis of a neural network trained to perform a syntax-sensitive task; we leave a more extensive investigation for future work.

\begin{table*}[ht]
\centering
\scalebox{0.75}{
\begin{tabular}{lllll}
\toprule
Training objective & Sample input & Training signal & Prediction task & Correct answer \\
\toprule
Number prediction & \textit{The keys to the cabinet} & \textsc{plural} & \textsc{singular/plural?} & \textsc{plural}\\
Verb inflection & \textit{The keys to the cabinet [is/are]} & \textsc{plural}     &  \textsc{singular/plural?} & \textsc{plural}\\
Grammaticality & \textit{The keys to the cabinet are here.} & \textsc{grammatical} & \textsc{grammatical/ungrammatical?} & \textsc{grammatical} \\
Language model & \textit{The keys to the cabinet} & are & $P(\textit{are}) > P(\textit{is})$? & True\\
\bottomrule
\end{tabular}
}
\caption{\label{table:task_examples}Examples of the four training objectives and corresponding prediction tasks.}
\end{table*}

\section{\label{sec:alternative}Alternative Training Objectives}

The number prediction task followed a fully supervised objective, in which the network identifies the
number of an upcoming verb based only on the words preceding the verb. This section proposes three objectives that modify some of the goals and assumptions of the number prediction objective (see Table \ref{table:task_examples} for an overview).

\paragraph{Verb inflection:}

This objective is similar to number prediction, with one difference:
the network receives not only the words leading up to the verb, but also the
singular form of the upcoming verb (e.g., \textit{writes}). In practice, then,
the network needs to decide between the singular and plural forms of a
particular verb (\textit{writes} or \textit{write}).  Having access to the
semantics of the verb can help the network identify the noun that serves as its
subject without using the syntactic subjecthood criteria. For example, in the following
sentence:

\ex. People from the capital often eat pizza.

only \textit{people} is a plausible subject for \textit{eat}; the network can
use this information to infer that the correct form of the verb is \textit{eat} is rather
than \textit{eats}. 

This objective is similar to the task that humans face during language
production: after the speaker has decided to use a particular verb
(e.g., \textit{write}), he or she needs to decide whether its form
will be \textit{write} or \textit{writes} \cite{levelt1999theory,staub2009interpretation}.

\paragraph{Grammaticality judgments:}

The previous objectives explicitly indicate the location in the
sentence in which a verb can appear, giving the
network a cue to syntactic clause boundaries. They also explicitly direct the network's attention to the number of the verb. As a form of weaker supervision,
we experimented with a grammaticality judgment objective. In this scenario, the network
is given a complete sentence, and is asked to judge whether or not it is grammatical. 
\begin{figure*}
    \raggedright
    \sidesubfloat[]{
        \includegraphics[width=3in]{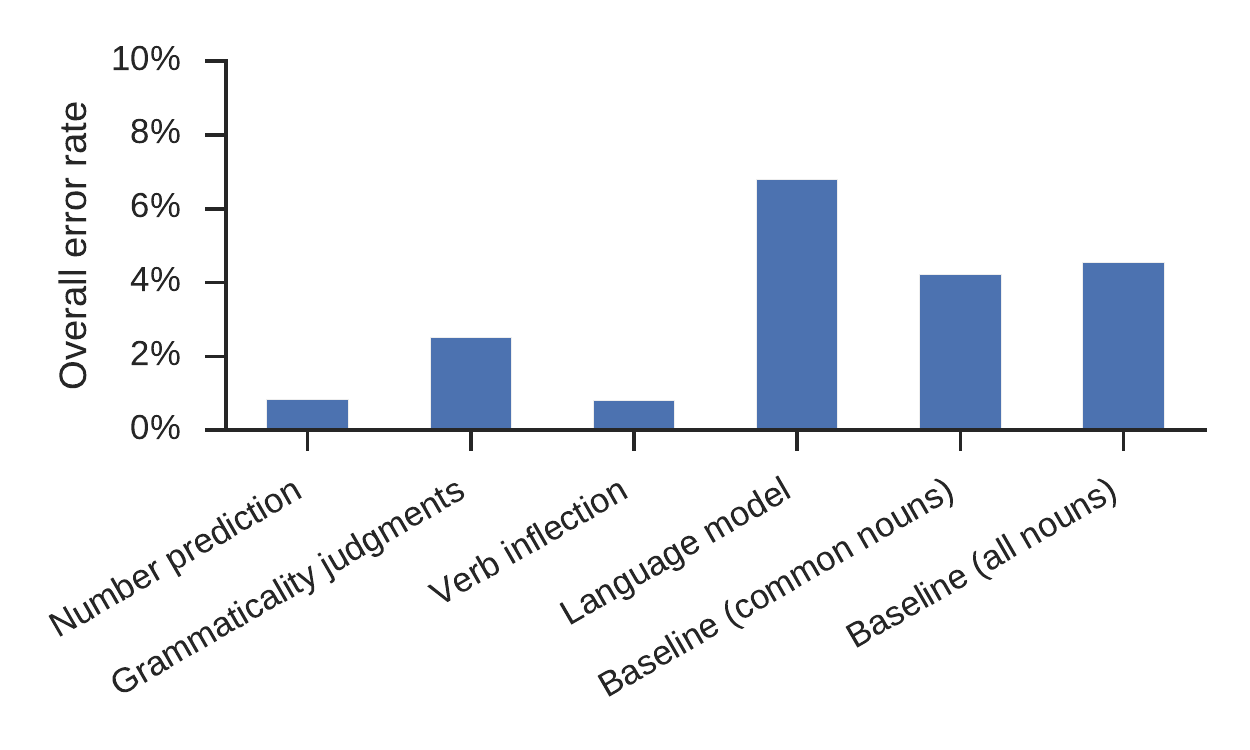}
        \label{fig:overall_across_tasks}
    }
    \sidesubfloat[]{
        \includegraphics[width=2.7in]{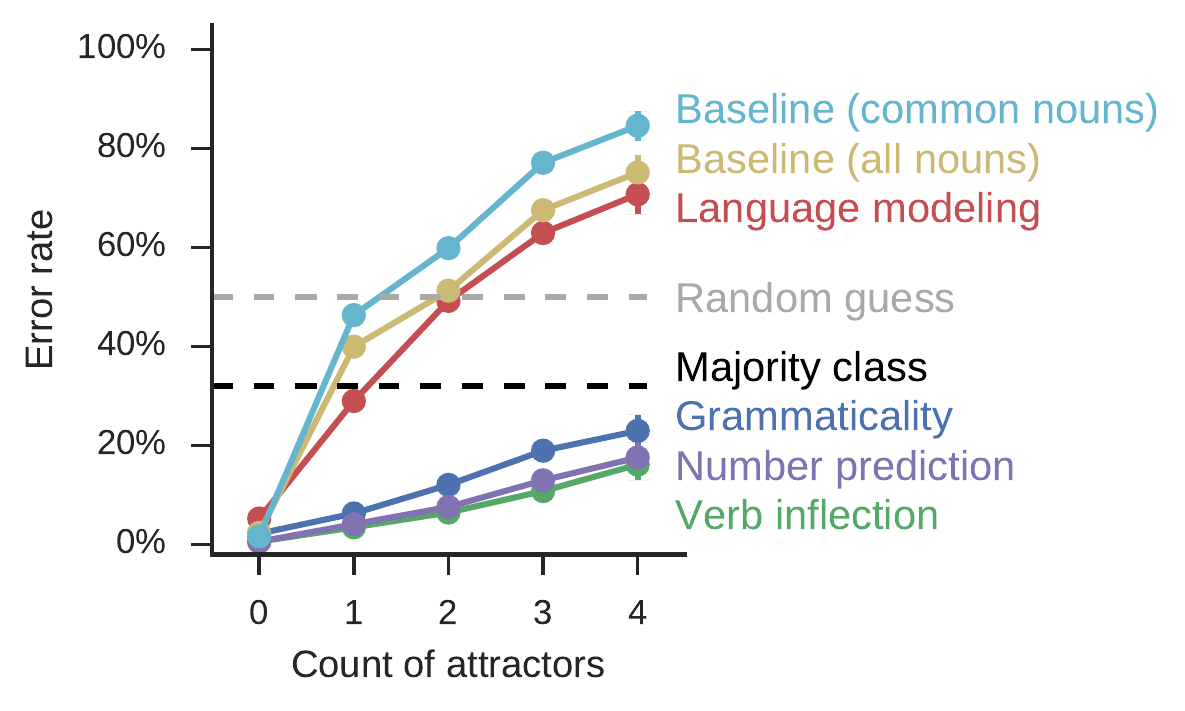}
        \label{fig:main_tasks}
    }
    
        \sidesubfloat[]{
        \vspace{0.5in}
        \includegraphics[width=1.3in, trim={0.2in 0 0.4in 0in}]{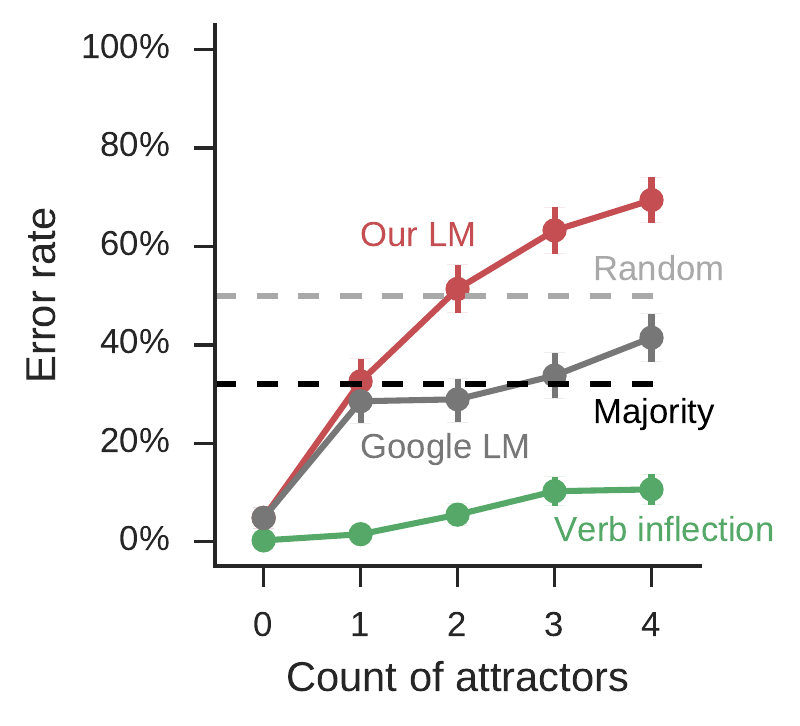}
   \label{fig:lms}}
    \sidesubfloat[]{
        \includegraphics[width=1.5in, trim={0.2in 0 0.4in 0}]{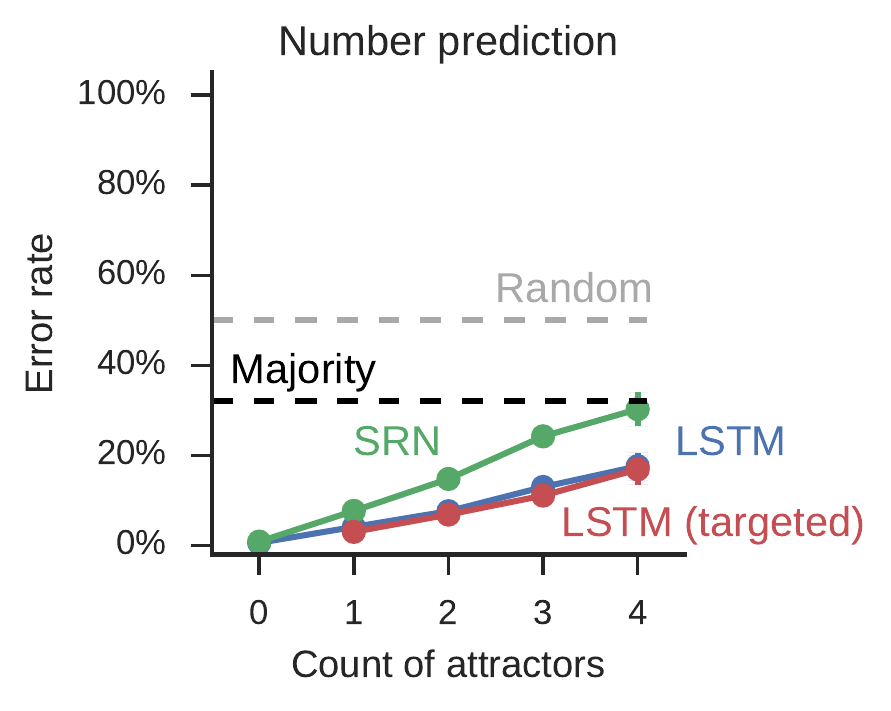}    \label{fig:number_prediction_comparison}
    }
    \sidesubfloat[]{
  \label{fig:subject_number_srn}.
    \includegraphics[width=2.7in, trim={0.2in 0 0.4in 0.1in}]{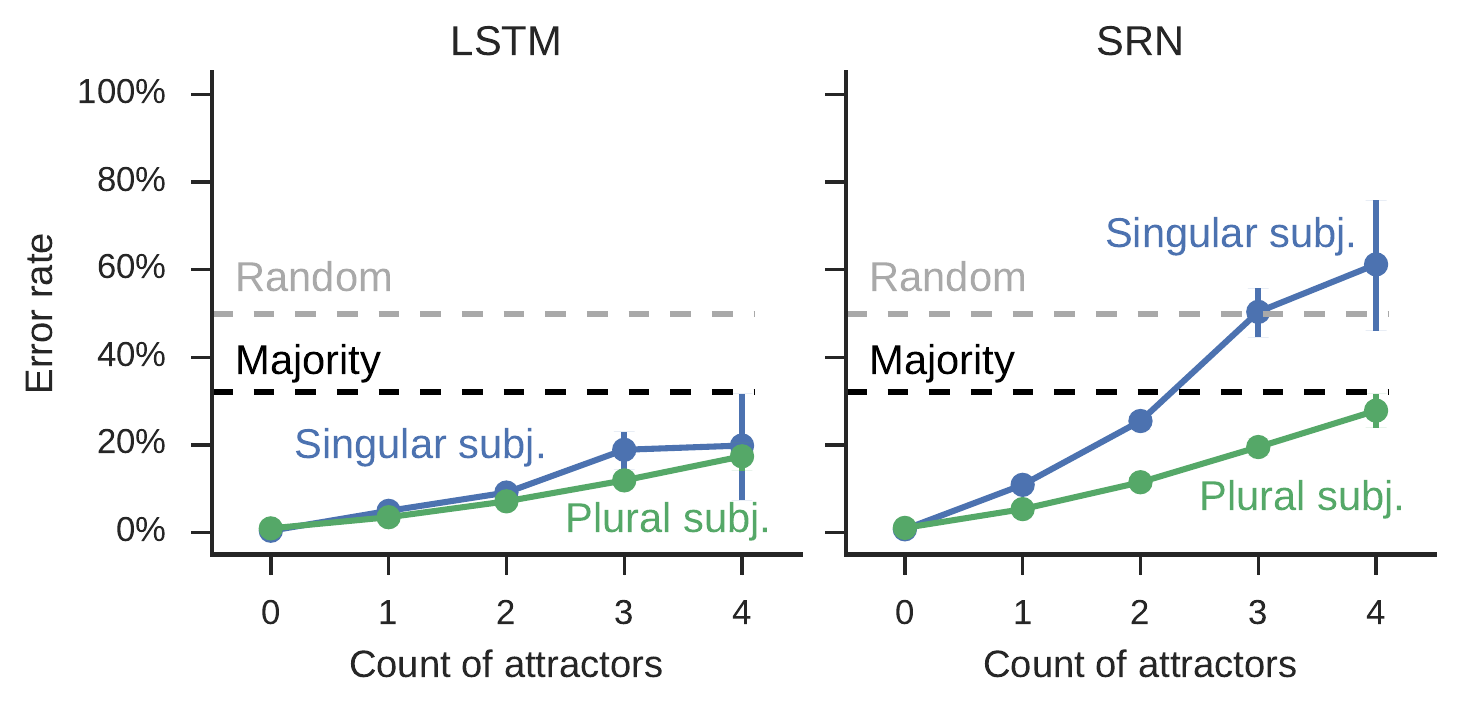}
    }
    \vspace{-0.2cm}

    \caption{Alternative tasks and additional experiments:
    \protect\subref{fig:overall_across_tasks} overall error rate across tasks
    (note that the y-axis ends in 10\%);
    \protect\subref{fig:main_tasks} effect of count of attractors in homogeneous
    dependencies across training objectives; 
    \protect\subref{fig:lms} comparison of the Google LM \protect\cite{jozefowicz2016exploring} to our LM and one of our supervised verb inflection systems, on a sample of sentences;
    \protect\subref{fig:number_prediction_comparison} number prediction: effect of count of
    attractors using SRNs with standard training or LSTM with targeted training;
    \protect\subref{fig:subject_number_srn} number prediction: difference in error rate between singular and plural subjects across RNN cell types. Error bars represent binomial 95\% confidence intervals.} 
    
\end{figure*}

To train the network, we made half of the examples in our training corpus
ungrammatical by flipping the number of the verb.\footnote{In some sentences
this will not in fact result in an ungrammatical sentence, e.g. with
collective nouns such as \textit{group}, which are compatible with both singular
and plural verbs
in some dialects of English \cite{huddleston2002cambridge}; those cases appear
to be rare.} The network read the entire sentence and received a supervision
signal at the end.  This task is modeled after a common human data collection
technique in linguistics \cite{schutze1996empirical}, although our training
regime is of course very different to the training that humans are exposed to:
humans rarely receive ungrammatical sentences labeled as such
\cite{bowerman1988no}.

\paragraph{Language modeling (LM):}

Finally, we experimented with a word prediction objective, in which the model
did not receive any grammatically relevant supervision
\cite{elman1990finding,elman1991distributed}. In this scenario, the goal of the
network is to predict the next word at each point in every sentence. It
receives unlabeled sentences and is not specifically instructed to attend to
the number of the verb. In the network that implements this training scenario,
RNN activation after each word is fed into a fully connected dense layer
followed by a softmax layer over the entire vocabulary.

We evaluate the knowledge that the network has acquired about subject-verb noun
agreement using a task similar to the verb inflection task. To perform the
task, we compare the probabilities that the model assigns to the two forms of
the verb that in fact occurred in the corpus (e.g., \textit{write} and
\textit{writes}), and select the form with the higher probability.%
\footnote{One could also imagine performing the equivalent of the number
prediction task by 
aggregating LM probability mass over all plural verbs and all singular verbs. This approach
may be more severely affected by part-of-speech ambiguous words than the one
we adopted; we leave the exploration of this approach to future work.}
As this task is not part of the network's training objective, and the model
needs to allocate considerable resources to predicting each word in the
sentence, we expect the LM to perform worse than the explicitly supervised objectives.

\paragraph{Results:}

When considering all agreement dependencies, all models achieved error rates 
below 7\% (Figure \ref{fig:overall_across_tasks}); as mentioned above, even
the noun-only number prediction baselines achieved error rates below 5\%
on this task. At the same time, there were large differences in accuracy
across training objectives.  The verb inflection network performed slightly but
significantly better than the number prediction one (0.8\% compared to
0.83\% errors), suggesting that the semantic information carried by the verb is moderately
  helpful. The grammaticality judgment objective performed somewhat worse, at
  2.5\% errors, but still outperformed the noun-only baselines by a large
  margin, showing the capacity of the LSTM architecture to learn syntactic
  dependencies even given fairly indirect evidence.

The worst performer was the language model. It made eight times as many errors
as the original number prediction network (6.78\% compared to 0.83\%), and did
substantially worse than the noun-only baselines (though recall that the
noun-only baselines were still explicitly trained to predict verb number). 

The differences across the networks are more striking when we focus on
dependencies with agreement attractors (Figure \ref{fig:main_tasks}). Here, the language model does worse
than chance in the most difficult cases, and only slightly better than the
noun-only baselines.  The worse-than-chance performance suggests that
attractors actively confuse the networks rather than cause them to make
a random decision.  The other models degrade more gracefully with the number of
agreement attractors; overall, the grammaticality judgment objective is
somewhat more difficult than the number prediction and verb inflection ones.
In summary, we conclude that while the LSTM is capable of learning
syntax-sensitive agreement dependencies under various objectives, the
language-modeling objective alone is not sufficient for learning such
dependencies, and a more direct form of training signal is required.

\paragraph{Comparison to a large-scale language model:} 

One objection to our language modeling result is that our LM faced a much
harder objective than our other models---predicting a distribution over 10,000
vocabulary items is certainly harder than binary classification---but was
equipped with the same capacity (50-dimensional hidden state and word vectors).
Would the performance gap between the LM and the explicitly supervised models
close if we increased the capacity of the LM?

We address this question using a very large publicly available LM
\cite{jozefowicz2016exploring}, which we refer to as the Google LM.\footnote{
\url{https://github.com/tensorflow/models/tree/master/lm_1b}} The Google LM
represent the current state-of-the-art in language modeling: it is trained on
a billion-word corpus \cite{chelba2013one}, with a vocabulary of 800,000 words.
It is based on a two-layer LSTM with 8192 units in each layer, or more than 300
times as many units as our LM; at 1.04 billion parameters it has almost 2000
times as many parameters. It is a fine-tuned language model that achieves
impressive perplexity scores on common benchmarks, requires a massive
infrastructure for training, and pushes the boundaries of what's feasible with
current hardware.

We tested the Google LM with the methodology we used to test ours.\footnote{One
technical exception was that we did not replace low-frequency words with their
part-of-speech, since the Google LM is a large-vocabulary language model, and
does not have parts-of-speech as part of its vocabulary.} Due to computational
resource limitations, we did not evaluate it on the entire test set, but
sampled a random selection of 500 sentences for each count of attractors
(testing a single sentence under the Google LM takes around 5 seconds on
average).  The results are presented in Figure \ref{fig:lms}, where they are
compared to the performance of the supervised verb inflection system. Despite
having an order of magnitude more parameters and significantly larger training
data, the Google LM performed poorly compared to the supervised models; even
a single attractor led to a sharp increase in error rate to 28.5\%, almost as
high as our small-scale LM (32.6\% on the same sentences). While additional
attractors caused milder degradation than in our LM, the performance of the
Google LM on sentences with four attractors was still worse than always
guessing the majority class (\textsc{singular}).

In summary, our experiments with the Google LM do not change our conclusions:
the contrast between the poor performance of the LMs and the strong performance
of the explicitly supervised objectives suggests that direct supervision has
a dramatic effect on the model's ability to learn syntax-sensitive
dependencies. Given that the Google LM was already trained on several hundred
times more data than the number prediction system, it appears unlikely that its
relatively poor performance was due to lack of training data.

\section{\label{sec:additional_experiments}Additional Experiments}

\paragraph{Comparison to simple recurrent networks:}

How much of the success of the network is due to the LSTM cells?  We repeated
the number prediction experiment with a simple recurrent network (SRN)
\cite{elman1990finding}, with the same number of hidden units.  The
SRN's performance was inferior to the LSTM's, but the average performance for
a given number of agreement attractors does not suggest a qualitative
difference between the cell types: the SRN makes about twice as many errors as the
LSTM across the board (Figure \ref{fig:number_prediction_comparison}).

\paragraph{Training only on difficult dependencies:}

Only a small proportion of the dependencies in the corpus had agreement
attractors (Figure \ref{fig:n_diff_intervening_count}).  Would the network
generalize better if dependencies with intervening nouns were emphasized during
training?  We repeated our number prediction experiment, this time training the
model only on dependencies with at least one intervening noun (of any number).
We doubled the proportion of training sentences to 20\%, since the total size
of the corpus was smaller (226K dependencies).

This training regime resulted in a 27\% decrease in error rate on dependencies
with exactly one attractor (from 4.1\% to 3.0\%). This decrease is
statistically significant, and encouraging given that total number of
dependencies in training was much lower, which complicates the learning of word
embeddings. Error rates mildly decreased in dependencies with more attractors
as well, suggesting some generalization (Figure
\ref{fig:number_prediction_comparison}). Surprisingly, a similar experiment
using the grammaticality judgment task led to a slight \textit{increase} in
error rate. While tentative at this point, these results suggest that
oversampling difficult training cases may be beneficial; a curriculum
progressing from easier to harder dependencies \cite{elman1993learning} may
provide additional gains.

\section{\label{sec:error_analysis}Error Analysis}

\paragraph{Singular vs. plural subjects:}

Most of the nouns in English are singular: in our corpus, the fraction of
singular subjects is 68\%. Agreement attraction errors in humans are much more
common when the attractor is plural than when it is singular
\cite{bock1991broken,eberhard2005making}. Do our models' error rates depend on
the number of the subject? 

As Figure \ref{fig:np_last_intervening} shows, our LSTM number prediction model
makes somewhat more agreement attraction errors with plural than with singular
attractors; the difference is statistically significant, but the asymmetry is
much less pronounced than in humans.  Interestingly, the SRN version of the
model does show a large asymmetry, especially as the count of attractors
increases; with four plural attractors the error rate reaches 60\% (Figure
\ref{fig:subject_number_srn}).

\paragraph{Qualitative analysis:}

We manually examined a sample of
200 cases in which the majority of the 20 runs of the number prediction network
    made the wrong prediction. There were only 8890 such dependencies (about
    0.6\%). Many of those were straightforward agreement attraction errors;
      others were difficult to interpret. We mention here three classes of
      errors that can motivate future experiments.

The networks often misidentified the heads of noun-noun compounds. In
\ref{ex:conservation}, for example, the models predict a singular verb even
though the number of the subject \textit{conservation refugees} should be
determined by its head \textit{refugees}. This suggests that the networks
didn't master the structure of English noun-noun compounds.\footnote{The
dependencies are presented as they appeared in the corpus; the predicted number
was the opposite of the correct one (e.g., singular in \ref{ex:conservation},
where the original is plural).}

\ex. \label{ex:conservation}Conservation \depmem{refugees} \depmem{live} in
a world colored in shades of gray; limbo.

\ex. \label{ex:it}Information technology (IT) \depmem{assets} commonly \depmem{hold} large
volumes of confidential data.

Some verbs that are ambiguous with plural nouns seem to have been misanalyzed
as plural nouns and consequently act as attractors. The models predicted
a plural verb in the following two sentences even though neither of them has
any plural nouns, possibly because of the ambiguous verbs \textit{drives} and
\textit{lands}:

\ex.The \depmem{ship} that the player drives \depmem{has} a very high speed.

\ex.It was also to be used to learn if the \depmem{area} where the lander lands \depmem{is} typical of the surrounding terrain.

Other errors appear to be due to difficulty not in identifying the subject but
in determining whether it is plural or singular. In Example
\ref{ex:five_paragraphs}, in particular, there is very little information in
the left context of the subject \textit{5 paragraphs} suggesting that the
writer considers it to be singular:

\ex.\label{ex:rabaul}Rabaul-based Japanese \depmem{aircraft} \depmem{make}
three dive-bombing attacks.

\ex.\label{ex:five_paragraphs}The lead is also rather long;
5 \depmem{paragraphs} \depmem{is} pretty lengthy for a 62 kilobyte article.

The last errors point to a limitation of the number prediction task, which jointly evaluates the model's ability to identify the subject and its ability to assign the correct number to noun phrases. 

\section{\label{sec:related_work}Related Work}

The majority of NLP work on neural networks evaluates them on their
performance in a task such as language modeling or machine translation \cite{sundermeyer2012lstm,bahdanau2015neural}. These evaluation setups average over many
different syntactic constructions, making it difficult to isolate the network's syntactic capabilities. 

Other studies
have tested the capabilities of RNNs to learn
simple artificial languages. \newcite{gers2001lstm} showed that LSTMs can learn the
context-free language $a^nb^n$, generalizing to $n$s as high as $1000$ even
when trained only on $n \in \{1, \ldots, 10\}$. Simple recurrent networks
struggled with this language \cite{rodriguez1999recurrent,rodriguez2001simple}.
These results have been recently replicated and extended by \newcite{joulin2015inferring}.

\newcite{elman1991distributed} tested an SRN on a miniature language that
simulated English relative clauses, and found that the network was only able to
learn the language under highly specific circumstances \cite{elman1993learning}, though later work has called some of his conclusions into question \cite{rohde1999language,cartling2008implicit}.
\newcite{frank2013acquisition} studied the acquisition of anaphora coreference
by SRNs, again in a miniature language. Recently, \newcite{bowman2015tree} tested the ability of LSTMs to learn an artificial language based on
propositional logic. As in our study, the performance of the network degraded as the complexity of the test sentences increased. 

\newcite{karpathy2016visualizing} present analyses and visualization methods
for character-level RNNs. \newcite{kadar2016representation} and
\newcite{li2016visualizing} suggest visualization techniques for word-level
RNNs trained to perform tasks that aren't explicitly syntactic (image
captioning and sentiment analysis).

Early work that used neural networks to model grammaticality judgments includes
\newcite{allen1999emergence} and \newcite{lawrence1996can}. More recently,
the connection between grammaticality judgments and the probabilities assigned by a language model
was explored by \newcite{clark2013statistical} and \newcite{lau2015unsupervised}. Finally, arguments for evaluating NLP models on a strategically sampled set of dependency types rather than a random sample of sentences have been made in the parsing literature \cite{rimell2009unbounded,nivre2010evaluation,bender2011parser}.

\section{\label{sec:discussion}Discussion and Future Work}

Neural network architectures are typically evaluated on random samples of
naturally occurring sentences, e.g., using perplexity on held-out data in
language modeling. Since the majority of natural language sentence are
grammatically simple, models can achieve high overall accuracy using flawed
heuristics that fail on harder cases. This makes it difficult to distinguish
simple but robust sequence models from more expressive architectures 
\cite{socher2014recursive,grefenstette2015learning,joulin2015inferring}. Our
work suggests an alternative strategy---evaluation on naturally occurring
sentences that are sampled based on their grammatical complexity---which can
provide more nuanced tests of language models
\cite{rimell2009unbounded,bender2011parser}.

This approach can be extended to the training stage: neural networks can be encouraged to develop more sophisticated generalizations by oversampling grammatically challenging training sentences. We took a first step in this direction when we trained the network only on dependencies with intervening nouns (Section \ref{sec:additional_experiments}). This training regime indeed improved the performance of the network; however, the improvement was quantitative rather than qualitative: there was limited generalization to dependencies that were even more difficult than those encountered in training. Further experiments are needed to establish the efficacy of this method.

A network that has acquired syntactic representations sophisticated enough to
handle subject-verb agreement is likely to show improved performance on other
structure-sensitive dependencies, including pronoun coreference, quantifier
scope and negative polarity items. As such, neural models used in NLP
applications may benefit from grammatically sophisticated sentence
representations developed in a multitask learning setup
\cite{caruana1998multitask}, where the model is trained concurrently on the
task of interest and on one of the tasks we proposed in this paper. Of course,
grammatical phenomena differ from each other in many ways. The distribution of
negative polarity items is highly sensitive to semantic factors
\cite{giannakidou2011negative}. Restrictions on unbounded
dependencies \cite{ross1967constraints} may require richer syntactic
representations than those required for subject-verb dependencies. 
The extent to which the results of our study will generalize to other
constructions and other languages, then, is a matter for empirical research. 

Humans occasionally make agreement attraction mistakes during language
production \cite{bock1991broken} and comprehension \cite{nicol1997subject}.
These errors persist in human acceptability judgments \cite{tanner2014time},
which parallel our grammaticality judgment task. Cases of grammatical agreement
with the nearest rather than structurally relevant constituent have been
documented in languages such as Slovenian \cite{maruvsivc2007last}, and have
even been argued to be occasionally grammatical in English
\cite{zwicky2005agreement}. In future work, exploring the relationship between
these cases and neural network predictions can shed light on the cognitive
plausibility of those networks. 

\section{\label{sec:conclusion}Conclusion}

LSTMs are sequence models; they do not have built-in hierarchical
representations.  We have investigated how well they can learn subject-verb
agreement, a phenomenon that crucially depends on hierarchical syntactic
structure.  When provided explicit supervision, LSTMs were able to learn to
perform the verb-number agreement task in most cases, although their error rate
increased on particularly difficult sentences. We conclude that LSTMs can learn
to approximate structure-sensitive dependencies fairly well given explicit
supervision, but more expressive architectures may be necessary to eliminate
errors altogether.  Finally, our results provide evidence that the language
modeling objective is not by itself sufficient for learning structure-sensitive
dependencies, and suggest that a joint training objective can be used to
supplement language models on tasks for which syntax-sensitive dependencies are
important.

\section*{Acknowledgments}
We thank Marco Baroni, Grzegorz Chrupa\l{}a, Alexander Clark, Sol Lago, Paul Smolensky, Benjamin Spector and Roberto Zamparelli for comments and discussion. This research was supported by the European Research Council (grant ERC-2011-AdG 295810 BOOTPHON), the Agence Nationale pour la   
Recherche (grants ANR-10-IDEX-0001-02 PSL and ANR-10-LABX-0087 IEC) and the Israeli Science Foundation (grant number 1555/15).

\begin{figure*}[h]
\includegraphics[width=\linewidth]{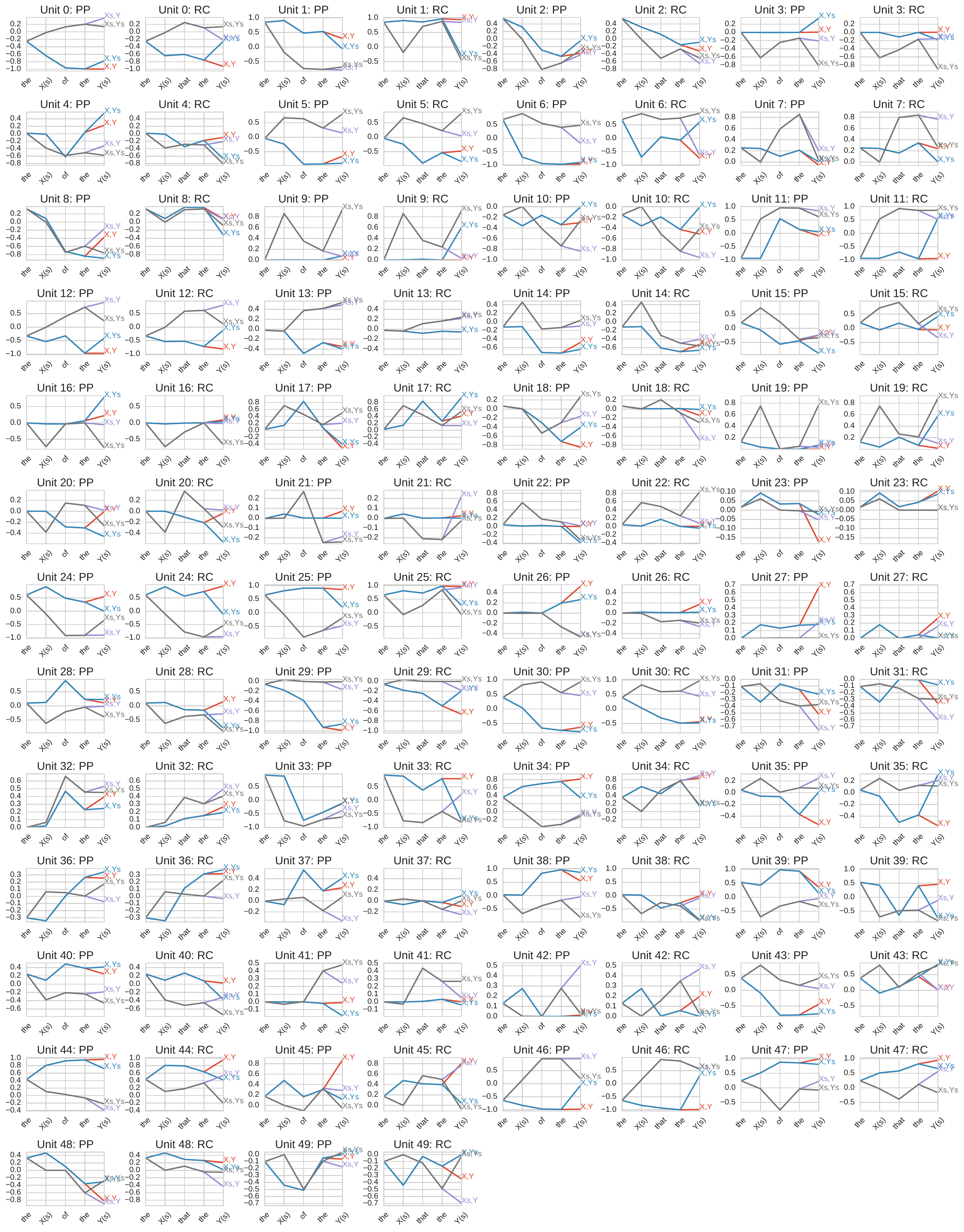}
\caption{\label{supplemental}Activation plots for all units (see Figure \ref{fig:lstmviz_factorial}
and text in p. 7).}
\end{figure*}

\bibliographystyle{acl2012}
\bibliography{bibexport.bib}
\end{document}